\documentclass{article}
\usepackage{amsfonts, amsmath, amsthm, amssymb}

\usepackage{authblk}
\usepackage{graphicx}
\usepackage{subfigure}
\usepackage{hyperref}

\theoremstyle{definition}
\numberwithin{equation}{section}

\begin{document}

\title{Centerpoints Are All You Need in Overhead Imagery}

\author[1]{James Inder}
\author[2]{Mark Lowell\thanks{Corresponding author: Mark.C.Lowell@nga.mil}\footnote{Equal contribution}}
\newcommand\CoAuthorMark{\footnotemark[\arabic{footnote}]}
\author[2]{A.J. Maltenfort\protect\CoAuthorMark}


\affil[1]{Booz Allen Hamilton}
\affil[2]{National Geospatial-Intelligence Agency}

\maketitle

\begin{abstract}

	Labeling data to use for training object detectors is expensive and time consuming. Publicly available overhead datasets for object detection are labeled with image-aligned bounding boxes, object-aligned bounding boxes, or object masks, but it is not clear whether such detailed labeling is necessary. To test the idea, we developed novel single- and two-stage network architectures that use centerpoints for labeling. In this paper we show that these architectures achieve nearly equivalent performance to approaches using more detailed labeling on three overhead object detection datasets.

\end{abstract}

\section{Introduction}

Every day, observation satellites capture terabytes of imagery of the Earth's surface that feed into a wide variety of civil and military applications. This stream of data has grown so large that only automated methods can feasibly analyze it. One critical component of remote sensing analysis is object detection: locating objects of interest on the Earth's surface in overhead imagery. Automated object detection algorithms have advanced by leaps and bounds over the last decade, but they still require vast amounts of labeled data for training, which is expensive and tedious to produce. Any technique that can reduce the resources needed to label objects in overhead imagery is therefore desirable.

Most existing datasets for training overhead object detectors are labeled with horizontal bounding boxes \cite{xview}\cite{vhr-10}\cite{rsod}\cite{dior}\cite{simd}, object-aligned bounding boxes \cite{dota}\cite{fair1m}\cite{hrsc2016}\cite{ucas-aod}\cite{vedai}, or segmentation masks \cite{potsdam}\cite{isaid}. These methods of labeling appear to have been inherited from work on natural images -- primarily cell phone pictures. Unlike objects in cell phone pictures, objects in overhead images are only seen in a narrow range of viewpoints and scales. This paper examines whether the extra work required to create such detailed labels is worthwhile in terms of the resulting detector performance.

In this paper, we show that centerpoints alone are sufficient for training overhead object detectors for most targets in overhead imagery and that they require significantly less time and work by labelers then image-aligned or object-aligned bounding boxes. We designed single- and two-stage object detection architectures for centerpoints based on RetinaNet \cite{retinanet} and Faster Region-Based Convolutional Neural Network (Faster R-CNN) \cite{faster-rcnn}. We compare the performance of our Centerpoint RetinaNet and Centerpoint R-CNN against RetinaNet and Faster R-CNN trained with horizontal and object-aligned bounding boxes on a variety of overhead datasets, and show that our centerpoint detectors match or exceed the performance of bounding box detectors.


In Section~\ref{sec:related}, we review past work on object detection, focusing on overhead imagery. In Section~\ref{sec:methodology}, we describe our centerpoint architectures and our methods for evaluating detectors for centerpoints, horizontal bounding boxes, and object-aligned bounding boxes on a common basis. In Section~\ref{sec:experiments}, we present the results of our experiments using each detector on a variety of overhead datasets. In Section~\ref{sec:conclusion}, we conclude by discussing the implications of our results for further work in object detection in overhead imagery.

\section{Related Work}
\label{sec:related}

\textbf{Labeling Methods in Overhead Imagery Datasets:} A survey of overhead object detection datasets shows that most use horizontal bounding boxes \cite{xview}\cite{vhr-10}\cite{rsod}\cite{dior}\cite{simd}, object-aligned bounding boxes \cite{dota}\cite{fair1m}\cite{hrsc2016}\cite{ucas-aod}\cite{vedai}, or segmentation masks \cite{potsdam}\cite{isaid}. Although the cost and difficulty of labeling large object detection datasets is generally acknowledged, regardless of the domain, we are unaware of any published systematic studies of the costs and benefits of different labeling approaches. Published efforts to reduce labeling costs for overhead imagery have instead focused on the use of synthetic data \cite{simpl}\cite{rareplanes}\cite{synthinel}\cite{synteo}. Outside of the overhead domain specifically, approaches include active learning \cite{settles}\cite{al-survey}, weak supervision \cite{weak-supervision}, few-shot learning \cite{few-shot-survey}, zero-shot learning \cite{zero-shot-survey}, and semi-supervised learning \cite{self-supervised-survey}\cite{semi-supervised-survey}. However, networks trained solely on synthetic imagery struggle to match the performance of networks trained with real, fully annotated data, and the other approaches all require at least some human annotation.

A small number of works have examined the use of point annotations. Papadopoulous et al. 2017 \cite{click-supervision} used Amazon Mechanical Turk to relabel the PASCAL VOC object detection dataset with centerpoints and then used those centerpoints to train object detectors. They found that nearly equivalent accuracy could be obtained at substantially lower labeling cost. However, instead of training a detector to predict centerpoints, they used the Edge-Boxes algorithm \cite{edge-boxes} to propose bounding boxes for the centerpoints and trained a Fast R-CNN detector \cite{fast-rcnn} to classify the proposals. Fast R-CNN is now obsolete compared to networks that generate their own proposals, such as Faster R-CNN \cite{faster-rcnn}, which combine higher performance with a faster runtime.

Mundhenk et al. 2016 \cite{cars} labeled cars in overhead imagery using centerpoints and trained sliding window classifiers and regression networks to count them in aerial images. They experimented with object detection using a heatmap approach with a strided classifier, but they did not compare their performance to networks trained with bounding box labels.

The work closest to our own is Ribera et al. 2019 \cite{locating-without-boxes}, which labeled computer vision datasets using centerpoints, including a dataset of overhead imagery, and trained a modified U-Net \cite{u-net} to predict those centerpoints using Hausdorff distance. They showed that their U-Net achieved equal or superior performance to a Faster R-CNN that predicted bounding boxes, but the Faster R-CNN used a different feature extractor architecture and was trained by imputing fixed-size bounding boxes to the centerpoints. They did not address whether the Faster R-CNN would have performed better if it had been trained with true bounding boxes tight around the targets or whether the difference in performance was attributable to the architecture of the feature extractor. Deshapriya et al. 2021 \cite{centroid-unet} trained a similar network using Gaussian kernels on a dataset of buildings and a dataset of coconut trees but did not compare their results to conventional object detectors.

\textbf{Object Detectors:} Modern object detectors can be classified as single-stage or multi-stage. Single-stage detectors such as RetinaNet \cite{retinanet} treat object detection as a regression problem. They use a backbone such as a ResNet \cite{resnet} as a feature extractor and then pass these features through a region proposal network to produce a set of predictions. Each prediction consists of class logits and offsets to an associated anchor box. These predictions are then compared to the ground truth and trained directly using a regression loss.

Multi-stage detectors such as Faster R-CNN \cite{faster-rcnn} follow the region proposal network by subsampling the proposals, then using a ROIAlign operation to crop features corresponding to the proposals out of the features from the backbone. These features are then passed to a classifier head, which predicts both the class logits and a set of corrections to the anchor box offsets. Some multi-stage networks such as ROI Transformer \cite{roi-transformer} repeat this process several times, refining the prediction at each stage. Multi-stage methods tend to perform slightly better than single-stage methods in public rankings, but they are significantly slower.

All of these detectors make their predictions as bounding boxes. These are usually horizontal bounding boxes, but variants using object-aligned boxes or segmentation masks have been created for both single-stage and multi-stage detectors \cite{mask-rcnn}\cite{roi-transformer}\cite{rotated-faster-rcnn}\cite{odtk}. Some detectors incorporate a centerpoint prediction, most famously Duan et al. 2019 \cite{center-net}, but only as a step in predicting a bounding box. The only work that we are aware of on detectors that specifically predict centerpoints is Ribera et al. 2019 \cite{locating-without-boxes}, but this work cannot be directly compared to existing bounding box detectors because of the difference in feature extractor architecture. Regression networks have been trained to generate a heatmap as part of their processing \cite{cars}\cite{buildings}\cite{buildings2}, but this heatmap is used to generate a count of targets in the image rather than to localize centerpoints. Kuzin et al. 2021 \cite{damaged-buildings} used single points as supervision in training an object detector for damaged buildings in overhead images of disaster areas, but the points were used only for the classification component, with the detection component trained conventionally on a different dataset. Finally, some studies of weak supervision, e.g. Bearman et al. 2016 \cite{whats-the-point}, have used single-point annotations for training, but these approaches are still judged on how well they perform segmentation, not on object detection.

\section{Methodology}
\label{sec:methodology}

\subsection{Centerpoint RetinaNet and Centerpoint R-CNN}

Adapting a single-stage object detector such as RetinaNet \cite{retinanet} to centerpoint detection is straightforward. Instead of predicting the center offset, height offset, and width offset for each anchor, we predict only the center offset. We assign ground truth to proposals based on Euclidean distance instead of intersection-over-union and train the regression loss using the smooth $L^1$ loss \cite{fast-rcnn}.

However, most state-of-the-art results in object detection are achieved by multistage object detectors such as Faster R-CNN \cite{faster-rcnn} and ROI Transformer \cite{roi-transformer}. These detectors consist of at least three components: a backbone, a region proposal network, and a classifier head. The backbone generates a set of features from the image, which the region proposal network uses to generate proposed bounding boxes. The bounding boxes may be horizontal or object-aligned. The proposals are subsampled, and features corresponding to each selected proposal are extracted by a pooling operation. The classifier head then classifies the proposals using the extracted features. However, if we are given only centerpoints, we have no bounding box with which to extract the features for the proposal.

Our Centerpoint R-CNN is a two-stage detector based on the Faster R-CNN. As with the Centerpoint RetinaNet, we assign ground truth to proposals using Euclidean distances and use the smooth $L^1$ loss instead of the intersection-over-union in the regression losses in both the region proposal network and the classifier head. To extract features for each proposal, we impute a fixed-size square bounding box to the detections for use in extracting the features for the proposals in the ROIAlign operation. We treat the box size as an additional hyperparameter.

\begin{figure*}
	\begin{center}
		\includegraphics[width=12.0cm]{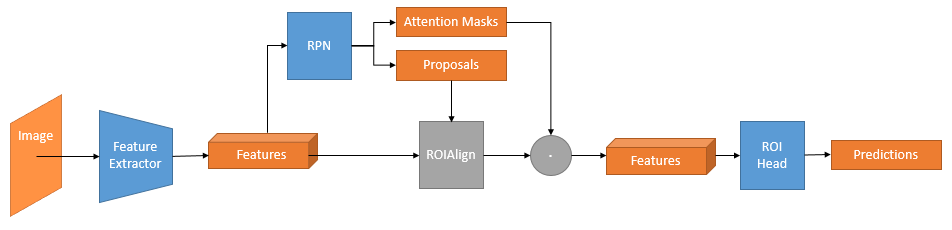}
	\end{center}
	\caption{Diagram of the Attentional RPN Mechanism}
	\label{fig:attentional-rpn}
\end{figure*}

Because of the fixed-size imputed bounding box, the proposals generated by the Region Proposal Network (RPN) in the Centerpoint R-CNN will tend to include significant amounts of background in addition to the target, which will confuse the classifier head in cluttered scenes. To handle this problem, we include an attention mechanism in our RPN, as shown in Figure~\ref{fig:attentional-rpn}. In addition to generating the proposal, the RPN outputs an attention mask with the same height and width as the proposal features extracted by the ROIAlign layer. The mask $m$ is produced by an additional $1\times1$ convolution on the end of the RPN, followed by a sigmoid function to rescale it to the range $[0, 1]$. It is then multiplied in a row-wise fashion by the features:
\[
	\hat{f}_{cxy} = m_{xy} f_{cxy}
\]
Here, $c$ indexes the channel; $x, y$ index the width and height; $f$ is the output of the ROIAlign layer; and $\hat{f}$ is the features passed to the classifier head. The mask predictor is trained by back-propagation from the classifier head, not by the RPN losses. Therefore, the mask predictor does \emph{not} predict a segmentation mask for the target, which would require segmentation annotations. Instead, the RPN learns to suppress areas within the imputed bounding box that would harm the performance of the classifier head, such as if the window area were to contain multiple distinct objects.

\subsection{Common Evaluation of Detectors Trained on Different Annotations}

To provide a consistent comparison between different forms of annotation, we converted all detections to centerpoints by taking the center of the bounding boxes. We then scored detections as true or false predictions based on the distance between the detection centerpoint and the target centerpoint. In datasets for which we have the ground sample distance (GSD) of all images, we used 3 meters as our cutoff for marking a detection as a true positive. In datasets for which we do not have GSD data for all images, we used 10 pixels as the cutoff. We calculated the average precision for each target class using this approach and then used the mean average precision across all classes as our evaluation metric.

\section{Experiments}
\label{sec:experiments}

\begin{figure*}
	\begin{center}
			\includegraphics[width=3.0cm]{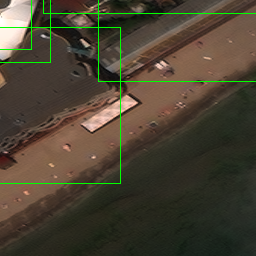}\hfill
			\includegraphics[width=3.0cm]{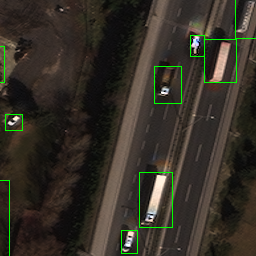}\hfill
			\includegraphics[width=3.0cm]{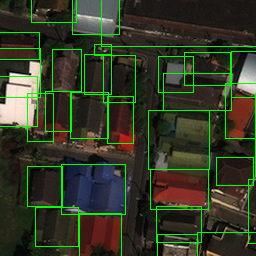}\hfill
			\includegraphics[width=3.0cm]{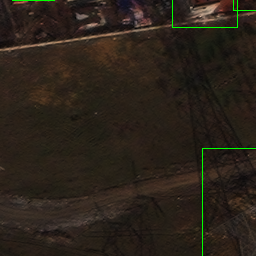}
			\\[\smallskipamount]
			\includegraphics[width=3.0cm]{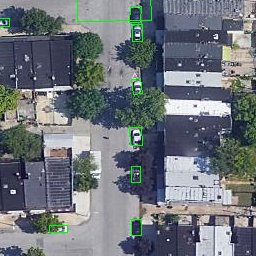}\hfill
			\includegraphics[width=3.0cm]{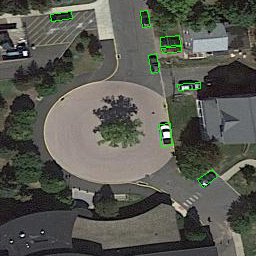}\hfill
			\includegraphics[width=3.0cm]{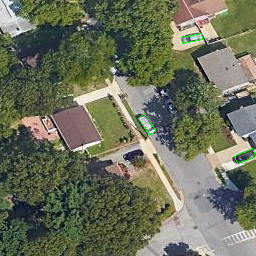}\hfill
			\includegraphics[width=3.0cm]{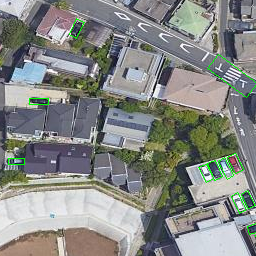}
			\\[\smallskipamount]
			\includegraphics[width=3.0cm]{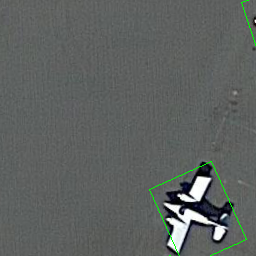}\hfill
			\includegraphics[width=3.0cm]{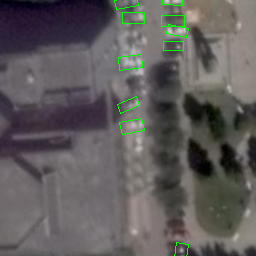}\hfill
			\includegraphics[width=3.0cm]{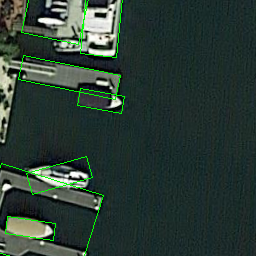}\hfill
			\includegraphics[width=3.0cm]{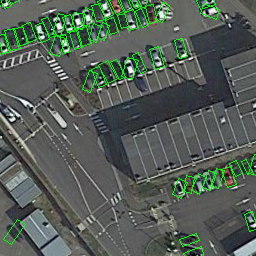}
	\end{center}
	\caption{Example Images - Top Row: xView; Middle Row: FAIR1M; Bottom Row: DOTA}
	\label{fig:examples}
\end{figure*}

We trained object detectors for centerpoints, horizontal bounding boxes, and object-aligned bounding boxes (when available) on the xView dataset \cite{xview}, DOTA 1.5 dataset \cite{dota}, and FAIR1M dataset \cite{fair1m}. Summaries of the datasets are shown in Table~\ref{table:datasets}. On the xView and FAIR1M datasets, we trained on the training dataset and tested on the validation dataset because the test datasets have not been publicly released. Examples from each dataset are shown in Figure~\ref{fig:examples}. We converted object-aligned boxes to image-aligned boxes by finding the tightest image-aligned box containing the object-aligned box. We converted boxes to centerpoints by taking the center of the box.

\begin{table*}[h!]
	\begin{center}
		\caption{Dataset Summaries: Note that our image and annotation counts include only the training and validation datasets for xView and FAIR1M because these were the only portions we used.}
		\label{table:datasets}
		\small
		\begin{tabular}{l|ccc|ccc}
			&&\textbf{\#}&\textbf{Image}&&\textbf{\#}&\textbf{\#}\\
			\textbf{Dataset}&\textbf{Labels}&\textbf{Classes}&\textbf{Size (pix)}&\textbf{Split}&\textbf{Images}&\textbf{Targets}\\
			\hline
			xView&Horizontal&60&$3,127\pm452$&Train&846&601,678\\
			     &          &  &             &Test &281&191,318\\
			DOTA 1.5&Rotated&18&$2,211\pm1,370$&Train&1,411&210,631\\
			        &              &  &               &Test &458&69,565\\
			FAIR1M&Rotated&37&$881\pm430$&Train&16,488&393,293\\
			      &              &  &           &Test&8,287&201,311
		\end{tabular}
	\end{center}
\end{table*}

We implemented our experiments using the detectron2 framework \cite{detectron2}. We used a Faster R-CNN and a RetinaNet for horizontal and object-aligned bounding boxes and our Centerpoint R-CNN and Centerpoint RetinaNet for centerpoints. We used a ResNet-101-FPN \cite{resnet}\cite{fpn} backbone for all experiments, initialized with a set of weights trained as a Faster R-CNN on the MS-COCO dataset \cite{ms-coco}. For all of our R-CNN networks, we expanded the pooler resolution to $14\times14$.

\begin{figure*}
	\begin{center}
		\includegraphics[width=12.0cm]{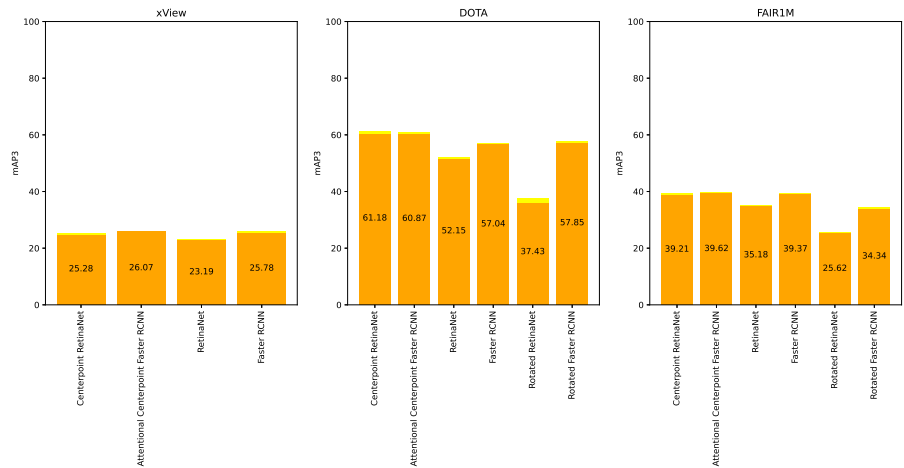}
		\caption{Performance of Detectors on Different Datasets\label{fig:results}}
	\end{center}
\end{figure*}

During training, we randomly sampled $800\times800$ chips from the datasets. We sampled 50 percent of our chips by selecting a random chip in a random image. We sampled the other 50 percent by randomly sampling a class, randomly sampling an example of that class, and randomly sampling a chip containing that example. For datasets for which we have GSDs, we randomly sampled a GSD between 0.1 m and 0.15 m and resized the image to that GSD before sampling the chip; for other datasets we applied a random resize between 66.7 percent and 150 percent. We additionally applied 90-degree random rotation, random horizontal flip, and color distortion as data augmentations. We trained for 90,000 iterations with a batch size of 16 and a base learning rate of 0.01, with 1,000 iterations of learning rate warmup, and $10\times$ learning rate shrinks at 60,000 and 80,000 iterations. We used mixed precision to reduce training runtime. A small number of runs failed to converge and were omitted from our results. The mean average precision for each network on each dataset is reported in Table~\ref{table:xView-results}, \ref{table:FAIR1M-results}, and \ref{table:DOTA-results}, and in Figure~\ref{fig:results}.

\begin{table*}[h!]
	\begin{center}
		\caption{Performance of Different Detectors on xView. mAP is mean average precision across classes, given as mean (standard deviation) across experiments. mAP -- S, mAP -- M, and mAP -- L are mAP on small, medium and large targets.}
		\label{table:xView-results}
		\small
		\begin{tabular}{lllll}
		\hline
		 Network          		     & mAP          & mAP -- S      & mAP -- M      & mAP -- L   \\
		\hline
		 Centerpoint RetinaNet 		     & 24.99 (0.29) & 15.32 (0.12) & 72.21 (1.67) & 40.98 (1.47) \\
		 Attentional Centerpoint             &              &              &              &              \\
		 Faster R-CNN                         & 26.03 (0.03) & 15.96 (0.37) & 63.24 (0.81) & 28.49 (1.93) \\
		 RetinaNet 		             & 22.93 (0.27) & 13.54 (0.12) & 59.62 (1.82) & 24.19 (2.28) \\
		 Faster R-CNN		             & 25.51 (0.27) & 15.25 (0.35) & 61.70 (0.77) & 21.31 (1.25) \\
		\hline
		\end{tabular}
	\end{center}
\end{table*}

\begin{table*}[h!]
	\begin{center}
		\caption{Performance of Different Detectors on FAIR1M. mAP is mean average precision across classes, given as mean (standard deviation) across experiments. mAP -- S, mAP -- M, and mAP -- L are mAP on small, medium and large targets.}
		\label{table:FAIR1M-results}
		\small
		\begin{tabular}{lllll}
		\hline
		 Network          		     & mAP          & mAP -- S      & mAP -- M      & mAP -- L   \\
		\hline
		 Centerpoint RetinaNet               & 38.90 (0.31) & 17.36 (0.20) & 80.81 (0.58) & 71.63 (3.00) \\
		 Attentional Centerpoint             &              &              &              &              \\
		 Faster R-CNN                         & 39.51 (0.12) & 18.08 (0.10) & 70.58 (0.80) & 60.00 (1.15) \\
		 RetinaNet                           & 35.06 (0.13) & 13.49 (0.08) & 68.27 (0.86) & 58.25 (1.81) \\
		 Faster R-CNN                         & 39.19 (0.17) & 18.28 (0.16) & 70.87 (1.24) & 55.79 (0.83) \\
 		 Rotated RetinaNet  		     & 25.51 (0.11) & 6.21 (0.08)  & 56.90 (0.50) & 59.70 (3.04) \\
		 Rotated Faster R-CNN   		     & 34.08 (0.26) & 11.35 (0.27) & 65.10 (0.70) & 58.89 (1.69) \\
		\hline
		\end{tabular}
	\end{center}
\end{table*}

\begin{table*}[h!]
	\begin{center}
		\caption{Performance of Different Detectors on DOTA 1.5. mAP is mean average precision across classes, given as mean (standard deviation) across experiments. mAP -- S, mAP -- M, and mAP -- L are mAP on small, medium and large targets.}
		\label{table:DOTA-results}
		\small
		\begin{tabular}{lllll}
		\hline
		 Network          		     & mAP          & mAP -- S      & mAP -- M      & mAP -- L    \\
		\hline
		 Centerpoint RetinaNet               & 60.68 (0.50) & 37.38 (0.54) & 86.87 (0.69) & 76.45 (1.70)  \\
		 Attentional Centerpoint             &              &              &              &               \\
		 Faster R-CNN                         & 60.53 (0.34) & 35.65 (0.52) & 81.79 (0.62) & 69.29 (0.56)  \\
		 RetinaNet                           & 51.73 (0.43) & 18.98 (0.25) & 81.64 (0.07) & 60.83 (2.71)  \\
		 Faster R-CNN                         & 56.89 (0.15) & 29.30 (0.36) & 82.70 (1.08) & 59.64 (0.25)  \\
		 Rotated RetinaNet 		     & 36.65 (0.78) & 7.47 (0.26)  & 74.20 (1.35) & 64.88 (1.19)  \\
 		 Rotated Faster R-CNN   		     & 57.36 (0.50) & 25.44 (0.89) & 83.34 (0.56) & 65.83 (1.13)  \\
		\hline
		\end{tabular}
	\end{center}
\end{table*}

\subsection{Cluttered Scenes}

One justification for more precise annotations such as object-aligned bounding boxes is that they should improve the network's performance in complex scenes, where the tighter bounding box reduces the amount of clutter in the proposal. To test this hypothesis, we binned the FAIR1M test images based on the ratio of the number of annotations to the number of pixels and calculated the mAP for each bin separately. The results are presented in Table~\ref{table:clutter}.

\begin{table*}[h!]
	\begin{center}
		\caption{Effect of Clutter on Performance. Each column gives mean average precision over the bin of images of the given percentiles of the ratio of annotations to number of pixels.}
		\label{table:clutter}
		\tiny
\begin{tabular}{lrrrrrrrrrr}
\hline
		       &   1\%- &  11\%- &  21\%- &  31\%- &  41\%- &  51\%- &  61\%- &  71\%- &  81\%- &  91\%-  \\
 Model                 &   10\% &   20\% &   30\% &   40\% &   50\% &   60\% &   70\% &   80\% &   90\% &   100\% \\
\hline
 Centerpoint RetinaNet &  35.3 &  42.2 &  39.1 &  42.7 &  40   &  37   &  43.9 &  44.6 &  32.6 &   29.1 \\
Attentional Centerpoint&       &       &       &       &       &       &       &       &       &        \\
 Faster R-CNN           &  36.7 &  42.3 &  41.8 &  44.7 &  41   &  37.8 &  45.3 &  44.9 &  32.9 &   29.6 \\
 RetinaNet             &  32.7 &  41.3 &  35.6 &  38.3 &  35.7 &  33.6 &  41   &  41   &  29.3 &   24.5 \\
 Faster R-CNN           &  36.5 &  43.9 &  38.8 &  42.9 &  38.9 &  37   &  44.5 &  44.5 &  33.4 &   30.5 \\
 Rotated RetinaNet     &  22.9 &  29.4 &  27.4 &  29.7 &  26.3 &  25.1 &  31   &  32.2 &  19.3 &   14.8 \\
 Rotated Faster R-CNN   &  33.2 &  39.1 &  36.2 &  38.8 &  35.9 &  34.8 &  40.8 &  41.9 &  28.8 &   22.7 \\
\hline
\end{tabular}
	\end{center}
\end{table*}

\subsection{Object Size and Pooler Window}

The Centerpoint R-CNN uses a fixed-size window for extracting the features for a proposal, where the size of the window is treated as a hyperparameter. It is natural to ask whether there is a correlation between the window size and performance against targets of a similar size --- e.g., is a larger window desirable against larger targets? We trained Centerpoint R-CNNs using a variety of window sizes on xView, and measured their performance against targets of different sizes, as shown in Table~\ref{table:object-size}. Interestingly, no strong relationship is apparent between window size and performance.

\begin{table*}[h!]
	\begin{center}
		\caption{Centerpoint R-CNN Window Size vs. Target Size. mAP is mean average precision across classes. mAP -- S, mAP -- M, and mAP -- L are mAP on small, medium and large targets.}
		\label{table:object-size}
		\begin{tabular}{rllllr}
		\hline
		   Window Size & mAP          & mAP -- S     & mAP -- M     & mAP -- L      \\
		\hline
		            20 & 25.60 (0.23) & 15.75 (0.11) & 62.04 (0.87) & 28.66 (1.64)  \\
		            35 & 25.41 (0.27) & 15.42 (0.38) & 62.12 (1.12) & 26.94 (1.94)  \\
		            70 & 25.54 (0.24) & 15.40 (0.29) & 62.29 (0.22) & 28.39 (1.86)  \\
		           100 & 25.93 (0.27) & 15.59 (0.34) & 61.81 (0.26) & 27.57 (0.96)  \\
		\hline
		\end{tabular}
	\end{center}
\end{table*}

\section{Conclusion}
\label{sec:conclusion}

Our work shows that full bounding boxes are not necessary to train effective object detectors for overhead imagery. More detailed annotations are still needed if the network needs to perform auxiliary tasks besides detection, such as determining the direction a ship is traveling. However, if the use case only involves detecting the presence of the target, not its size or orientation, then a single point is sufficient. This allows object detectors for new target classes to be created in less time for lower cost compared to traditional bounding box annotations.

\textbf{Acknowledgements:} This work was supported in part by high-performance computer time and resources from the DoD High Performance Computing Modernization Program.

Approved for public release, 22-701.

\end{document}